\documentclass[conference]{IEEEtran}
\IEEEoverridecommandlockouts

\usepackage{cite}
\usepackage{amsmath,amssymb,amsfonts}
\usepackage{algorithmic}
\usepackage{graphicx}
\usepackage{textcomp}
\usepackage{xcolor}
\usepackage{enumitem}
\usepackage{threeparttable}
\usepackage{makecell}
\usepackage{longtable}
\usepackage{multirow}
\usepackage[thicklines]{cancel}
\usepackage{hyperref}  %

\usepackage{rotating}
\usepackage{booktabs} 
\usepackage[hang]{footmisc}

\def\BibTeX{{\rm B\kern-.05em{\sc i\kern-.025em b}\kern-.08em
		T\kern-.1667em\lower.7ex\hbox{E}\kern-.125emX}}
\begin{document}
	
	\title{Learning to Retrieve for Environmental \\Knowledge Discovery: An Augmentation-Adaptive Self-Supervised Learning Framework
	}

\author{
    \IEEEauthorblockN{
        Shiyuan Luo\textsuperscript{1,†}, 
        Runlong Yu\textsuperscript{2,†}, 
        Chonghao Qiu\textsuperscript{1},
        Rahul Ghosh\textsuperscript{3}, \\
        Robert Ladwig\textsuperscript{4},
        Paul C. Hanson\textsuperscript{5}, 
        Yiqun Xie\textsuperscript{6}, 
        Xiaowei Jia\textsuperscript{1}
    }
    \IEEEauthorblockA{
        \textsuperscript{1}\textit{University of Pittsburgh, }
        \textsuperscript{2}\textit{University of Alabama, }
        \textsuperscript{3}\textit{University of Minnesota,} \\
        \textsuperscript{4}\textit{Aarhus University,}
        \textsuperscript{5}\textit{University of Wisconsin-Madison,} 
        \textsuperscript{6}\textit{University of Maryland} 
    }
   \IEEEauthorblockA{
   \{shl298,chq29,xiaowei\}@pitt.edu, ryu5@ua.edu, ghosh128@umn.edu, \\
    rladwig@ecos.au.dk, pchanson@wisc.edu,  xie@umd.edu
    }
    \thanks{\textsuperscript{†}Shiyuan Luo and Runlong Yu contributed equally to this research.}

    \vspace{-0.7cm}
}

\maketitle

\begin{abstract}
		The discovery of environmental knowledge depends on labeled task-specific data, but is often constrained by the high cost of data collection. Existing machine learning approaches usually struggle to generalize in data-sparse or atypical conditions. 
	To this end, we propose an \textit{Augmentation-Adaptive Self-Supervised Learning (A$^2$SL)} framework, which retrieves relevant observational samples to enhance modeling of the target ecosystem. Specifically, we introduce a multi-level pairwise learning loss to train a scenario encoder that captures varying degrees of similarity among scenarios. 
	These learned similarities drive a retrieval mechanism that supplements a target scenario with relevant data from different locations or time periods.
	Furthermore, to better handle variable scenarios, particularly under atypical or extreme conditions where traditional models struggle, we design an augmentation-adaptive mechanism that selectively enhances these scenarios through targeted data augmentation.  
	Using freshwater ecosystems as a case study, we evaluate A$^2$SL in modeling water temperature and dissolved oxygen dynamics in real-world lakes. Experimental results show that A$^2$SL significantly improves predictive accuracy and enhances robustness in data-scarce and atypical scenarios. Although this study focuses on freshwater ecosystems, the A$^2$SL framework offers a broadly applicable solution in various scientific domains.
\end{abstract}
\vspace{0.3em}
\noindent\textbf{Code}---\href{https://github.com/shiyuanlsy/A2SL}{https://github.com/shiyuanlsy/A2SL}


\section{Introduction}

Environmental issues and sustainable development rank among the most critical global challenges of the 21st century. Advancing environmental knowledge discovery requires extracting insights from complex ecological systems~\cite{yu2025environmental}, which relies on accurate and comprehensive data collection. However, environmental data collection is fraught with obstacles. The high costs associated with the deployment and maintenance of monitoring infrastructure, particularly in remote or inaccessible regions, and difficulties in data acquisition during specific periods (e.g., due to ice cover, cloud cover, or instrument failures) lead to substantial data sparsity and regional imbalances~\cite{willard2022integrating,hampton2013big,reichstein2019deep}. These pose considerable barriers to modeling and prediction efforts. 
	
Traditional process-based models are widely used to simulate environmental variables, grounding their predictions in fundamental physical principles such as mass and energy conservation. In freshwater ecosystems, for example, these models are commonly used to predict key water quality metrics, including water temperature and dissolved oxygen (DO) concentrations~\cite{hipsey2019general,ladwig2022long}, which are critical for assessing ecosystem health and ensuring water security. Similar approaches are also applied in fields such as geology, oceanography, atmospheric science, and climate research~\cite{reichstein2019deep}. 
Despite their wide applicability, process-based models face significant limitations. They often require extensive parameterization or depend on oversimplified approximations, primarily due to incomplete physical understanding and the inherent complexity of modeling intricate environmental processes~\cite{gupta2014debates,beven2006manifesto}. These limitations can introduce inherent biases that ultimately compromise the accuracy and reliability of the predictions.

Data-driven approaches, particularly those that leverage machine learning (ML), have gained increasing traction in environmental modeling~\cite{reichstein2019deep,willard2022integrating}. ML methods excel at capturing complex patterns and nonlinear relationships within data, often surpassing traditional models when sufficient high-quality data is available~\cite{karpatne2017theory}.
However, environmental ecosystems often exhibit significant spatio-temporal heterogeneity, while observation data are typically collected in an imbalanced manner across different locations and time periods due to the high cost of data acquisition~\cite{shen2018transdisciplinary}.
ML models are highly dependent on observation data. As a result, models trained on existing datasets often fail to generalize under atypical conditions, such as extreme weather events or rare ecological phenomena, due to overfitting or the lack of relevant training examples~\cite{lazer2014parable}.
	 
Recent advancements in generative AI have demonstrated the effectiveness of leveraging external information to augment context for target tasks~\cite{lewis2020retrieval,abootorabi2024multimodal,peng2024graphrag}. 
Such data augmentation has been largely enabled by advanced pre-trained models, which can retrieve relevant content from large-scale auxiliary corpus data. 
Inspired by this success,  we pose an important question: can environmental conditions similarly guide the retrieval of relevant data from different locations or time periods to supplement data-sparse scenarios? While promising, this approach presents two key technical challenges.

	The first challenge is to accurately quantify the similarity of different environmental scenarios to support the retrieval process.
	In NLP, the similarity or relevance of documents can be directly inferred by pre-trained models~\cite{devlin2018bert}, which are effectively trained on large and representative datasets using objectives such as masked token prediction.  
	However, these pre-training methods are not suitable for environmental modeling as they are not designed for capturing the complex dependencies between environmental input features (e.g., weather conditions) and target variables (e.g., ecosystem properties). Unlike structured text or images, environmental data is shaped by complex interactions among different variables and also contains substantial noise~\cite{yu2025survey,yu2025foundation}. 
	As a result, directly using these existing methods can lead to the retrieval of scenarios that either fail to preserve key scenario characteristics or introduce extraneous variations, which hinders the effectiveness in enhancing the target task.


	The second challenge lies in 
	effectively determining when and where to apply data augmentation across different environmental scenarios.  
	Environmental data contain both stable scenarios, which feature long-term and consistent trends,  and variable scenarios, which exhibit distinct patterns (e.g., significant fluctuations), particularly under atypical or extreme conditions~\cite{woolway2021phenological,appling2018overcoming}. 
	Although data augmentation could help traditional ML models better generalize to data-scarce variable scenarios, applying it indiscriminately over all scenarios may disrupt the consistency of stable environments, potentially degrading model performance. Striking the right balance requires a selective augmentation strategy to identify scenarios that are likely to benefit from  external knowledge 
	without distorting inherent consistent environmental dynamics~\cite{yu2024adaptive,ravirathinam2024combining}.


	To tackle these challenges, we propose an \textit{Augmentation-Adaptive Self-Supervised Learning (A$^2$SL)} framework. 
	The idea is to dynamically switch between specialized models for capturing stable and variable scenarios, while selectively retrieving and integrating relevant data sharing similar environmental patterns to augment the learning for data-sparse, variable scenarios. 
	In our framework, each scenario is defined by short-term environmental conditions, paired with simulated labels produced by process-based models. 
	We create two predictive models: one specialized for stable scenarios, ensuring consistent long-term predictions (e.g., yearly trends); and another tailored for variable scenarios, improving short-term predictions (e.g., monthly behaviors) under atypical or extreme conditions where data augmentation is most needed. 
	An augmentation-adaptive mechanism is proposed to determine which predictive model should be applied to a given scenario and whether augmentation is needed, ensuring an optimal prediction strategy for each scenario type. 
	To effectively retrieve relevant scenarios for augmentation, 
	we build a self-supervised scenario encoder that captures scenario similarities through a multi-level pairwise learning loss~\cite{yu2018multiple}. This loss guides the encoder in assessing the similarity of an anchor scenario with three types of other scenarios, i.e., positively augmented data scenarios, semi-positively augmented data scenarios, and negatively augmented data scenarios. 
	The learned representation and similarities enable a retrieval mechanism to gather relevant data from different locations or time periods, supplementing data-sparse variable scenarios. 
	Finally, a decoder is employed to integrate information from both the current and retrieved scenarios to predict target ecosystem variables, improving the model’s accuracy and robustness.

	We evaluate the proposed method using freshwater ecosystems as a case study, with a focus on modeling water temperature and DO dynamics in real-world lakes. These two variables are key indicators of water quality and ecosystem health~\cite{wilson2010water,solomon2013ecosystem,phillips2020time}. Water temperature regulates metabolic rates, chemical reactions, and oxygen solubility, shaping overall ecosystem dynamics~\cite{yu2025physics}. Similarly, DO concentrations are crucial for aquatic biodiversity and water safety. Oxygen fluctuations in lakes often reflect their ecological balance and overall health~\cite{birge1906gases}. Our evaluation across 356 lakes in the Midwestern United States demonstrates the effectiveness of the A$^2$SL framework in improving water temperature and DO predictions, even in data-sparse scenarios.
Our contributions are summarized as follows:
\begin{itemize}
	\item We propose a self-supervised learning approach that quantifies scenario similarity to retrieve relevant data across different locations and periods, enabling targeted retrieval and improving data availability in sparse settings.
	\item We develop an augmentation-adaptive mechanism that switches between models for stable and variable scenarios, improving robustness under diverse conditions.
	\item We validate A$^2$SL in 356 real-world freshwater lakes in the Midwestern United States, showing significant gains in predicting water temperature and DO dynamics.
\end{itemize}

\section{Problem Formulation}

\begin{figure} [!t]
	\centering
\includegraphics[width=0.87\linewidth]{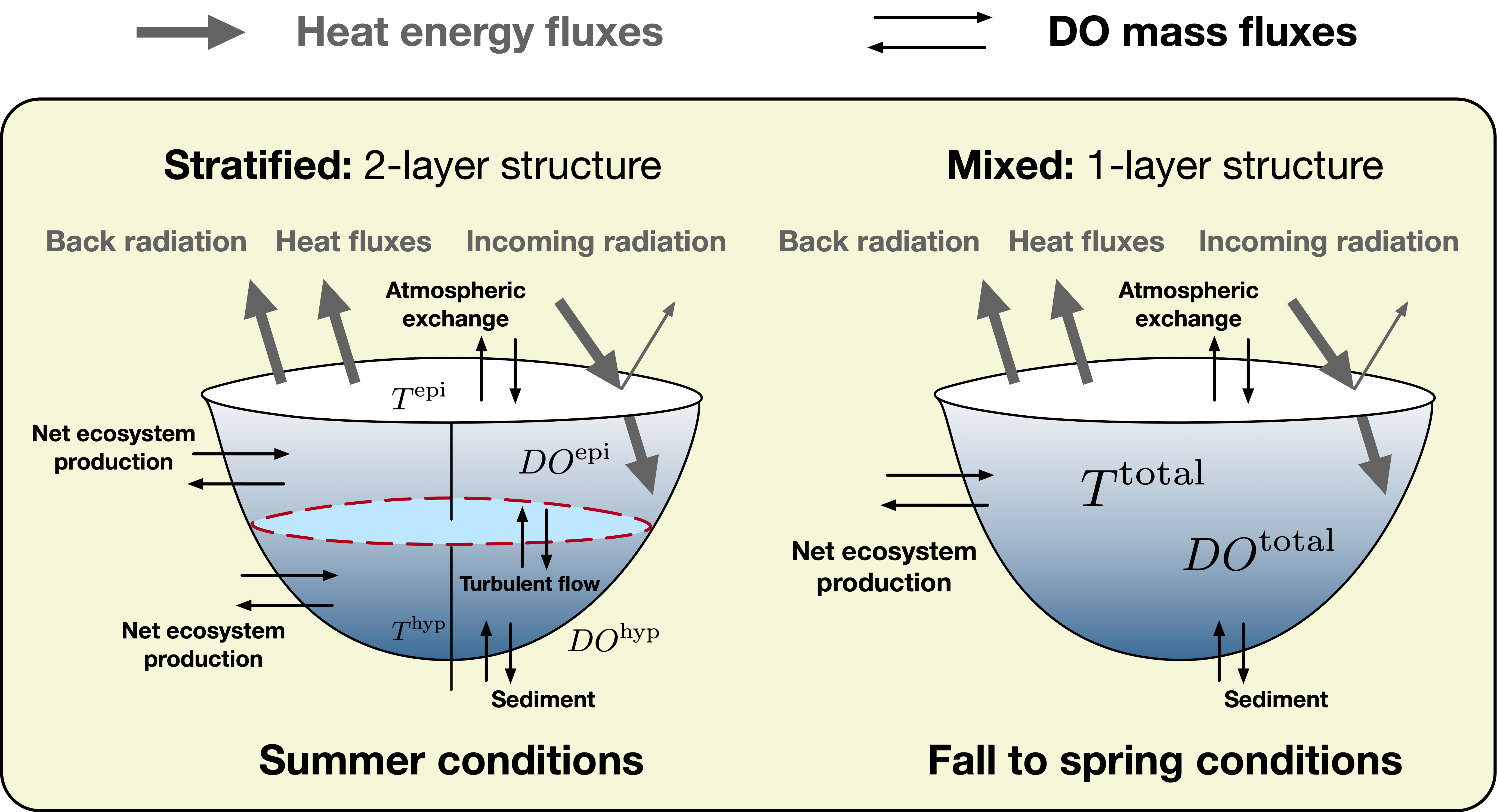}
\vspace{-0.1cm}
	\caption{Heat energy and DO mass fluxes in lake systems. } 
	\label{fig:1}
\end{figure}

The objective of this work is to predict daily water temperature and DO concentrations across numerous lake systems, spanning broad geographic regions and long time periods. As illustrated in Figure~\ref{fig:1}, water’s thermal expansion leads to stratification, forming a stable vertical density gradient. This stratification restricts vertical mixing, limiting nutrient and oxygen exchange between layers and reducing connectivity between bottom waters and the atmosphere, thereby hindering oxygen replenishment~\cite{read2011derivation}. During summer, stratified lakes exhibit a vertical density difference exceeding $0.05\,\text{kg/m}^3$ between surface and bottom layers, an average water temperature above $4\,^\circ\text{C}$, and a distinct thermocline. 

\begin{figure*} [!t]
	\centering
	\includegraphics[width=0.9\linewidth]{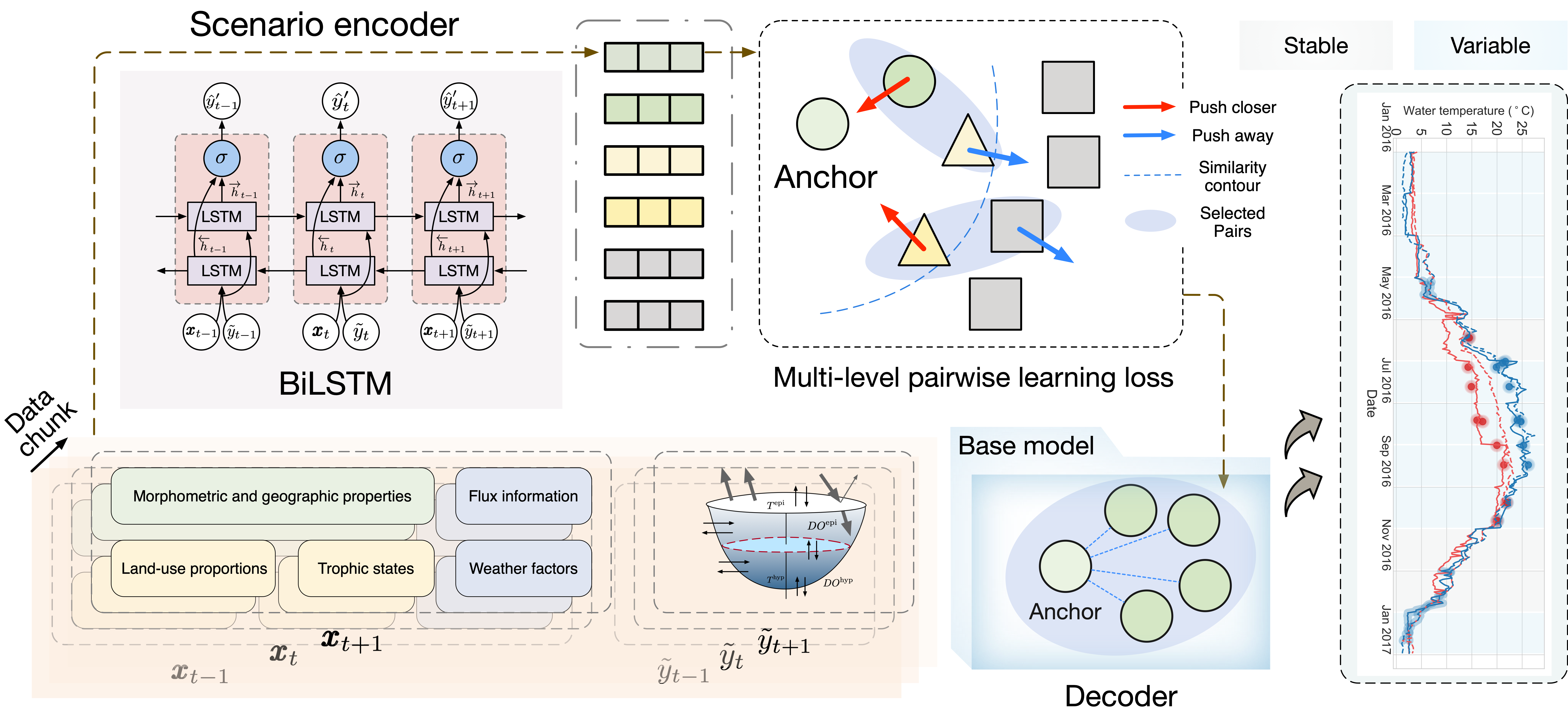}
    \vspace{-0.3cm}
	\caption{The overall framework of A$^2$SL. } 
	\vspace{-0.3cm}
	\label{fig:2}
\end{figure*} 

\textit{Water temperature prediction.}
During summer, we predict water temperature at two distinct depth layers: the epilimnion, a well-mixed surface layer, and the hypolimnion, a cooler, nutrient-rich, but light-limited deep layer. At each time step $t$, the epilimnion temperature is denoted as $T_t^{\rm epi}$, while the hypolimnion temperature is denoted as $T_t^{\rm hyp}$. From fall to spring, when the water column is typically well-mixed, we predict the overall water temperature of the entire lake, denoted as $T_t^{\rm total}$. 


\textit{DO concentration prediction.}
Similarly, during summer, DO concentrations are predicted separately for the epilimnion and hypolimnion at each time step $t$, denoted as $DO_{t}^{\rm epi}$ and $DO_{t}^{\rm hyp}$, respectively. 
This distinction accounts for the differences in oxygen dynamics and metabolic processes in each layer.
From fall to spring, when the water column undergoes full mixing, we predict the total DO concentration, $DO_{t}^{\rm total}$, representing the oxygen distribution across the entire lake.

For each target lake, we have access to its environmental features at each time $t$, represented as $\pmb{x}_{t} = \{x_{t}^1, \dots, x_{t}^m\}$, which influence lake temperature and DO. These features include morphometric and geographic properties such as lake area, depth, and shape; flux information such as ecosystem and sedimentation fluxes; weather factors including wind speed and air temperature; trophic states ranging from dystrophic to eutrophic; and land-use proportions, such as forests, wetlands, and other surrounding landscapes. In addition to these inputs, water temperature and DO concentrations are observed on a subset of days, denoted as $y_t$, where $y_t \in \{T_t^{\rm epi}, T_t^{\rm hyp}, T_t^{\rm total}, DO_{t}^{\rm epi}, DO_{t}^{\rm hyp}, DO_{t}^{\rm total}\}$. 
It is noteworthy that the observations are available only for a limited subset of time steps due to the cost of data collection. Also the number of observations can vary across different lakes.


\section{Methodology}

In this section, we present the \textit{Augmentation-Adaptive Self-Supervised Learning (A$^2$SL)} framework, as depicted in Fig.~\ref{fig:2}.
We begin with the scenario encoder, followed by the self-supervised learning approach for quantifying scenario similarity using a multi-level pairwise learning loss. The encoder is jointly trained with the decoder, which leverages these learned similarities to drive a retrieval mechanism, enriching sparse scenarios with relevant data from different locations or time periods. Finally, we introduce the augmentation-adaptive mechanism, which dynamically selects between two predictive models: one for stable scenarios and another for variable scenarios such as extreme or atypical conditions.

\setlength{\skip\footins}{6pt}

\subsection{Scenario Encoder}

\subsubsection{Scenario description} \label{sec:3A}
A \textit{scenario} $S$ is defined as a short-term sequence of environmental conditions that characterize the state of a specific environmental ecosystem. As illustrated in Fig.~\ref{fig:1}, lake dynamics arises from complex interactions among meteorological conditions, hydrological processes, and biochemical fluxes, and often exhibit varying patterns in short periods (e.g., certain months). Hence, we represent each scenario as a \textit{data chunk} $
S = \bigl\{(\pmb{x}_t, \tilde{y}_t)\bigr\}_{t=1}^n,
$ that captures the daily variations of 
relevant environmental conditions and flux variables in a lake over a one-month period.   
Here $\pmb{x}_t$ denotes observed environmental features and $\tilde{y}_t$ represents simulated outputs generated by process-based models, which capture general physical relationships driving the dynamics of target variables. The index $t$ represents each day within one month, forming a temporally ordered sequence\footnote{For consistency, we fix each month to 30 days (thus $n=30$ and 12 months per year); the remaining 5–6 calendar days are discarded.}. During training, the simulated series $\tilde{y}_t$ serves as an additional input feature.

For lake temperature simulation, we employ the General Lake Model (GLM)~\cite{hipsey2019general}, a widely adopted process-based model that captures key heat fluxes regulating water temperature. These fluxes include heat input from terrestrial long-wave and incoming short-wave radiation, as well as heat loss through back radiation, sensible heat flux, and latent evaporative heat flux. A portion of short-wave and long-wave radiation is reflected by the lake surface, further shaping thermal dynamics. For DO concentration, we utilize a process-based model following~\cite{ladwig2022long}, which simulates major DO mass fluxes, including atmospheric exchange, net ecosystem production, and oxygen consumption by sediment. During summer, it also accounts for DO entrainment fluxes driven by turbulent flow, thereby influencing thermocline depth and regulating oxygen transport between the epilimnion and hypolimnion.

\subsubsection{Self-supervised similarity learning}

We propose a self-supervised learning method to extract scenario representations that facilitate the retrieval of other scenarios governed by similar underlying environmental processes. 
Here we consider two scenarios to be \emph{physically consistent} if they differ only in a few dimensions (e.g., year, month, or lake identity) that do not fundamentally alter their underlying physical or environmental processes, e.g., the data of the same lake taken from different years but during the same calendar month. We aim to ensure that these scenarios under similar environmental processes remain close in the representation space learned through the self-supervised learning.

Formally, we perform a multi-level pairwise learning~\cite{yu2018multiple}, which captures the similarity relationships among an anchor scenario $S_A$, a \emph{positively} augmented scenario $S^+$, a \emph{semi-positively} augmented scenario $S^*$, and a \emph{negatively} augmented scenario $S^-$. 
To efficiently sample augmented scenarios, we employ a hierarchical sampling mechanism. First, we cluster all scenarios based on their morphometric and geographic properties. Within each cluster, we sample data chunks following predefined rules that ensure varying levels of physical and temporal consistency.
Intuitively, the positively augmented scenarios are selected from the closest clusters and are physically consistent with the anchor scenario $S_A$, but varies in only a few dimensions. These include scenarios from the same lake and month but from a different year, or from a different lake that shares the same month and year with $S_A$. 
The semi-positively augmented scenarios come from relatively close clusters and deviate more significantly in temporal patterns while remaining largely consistent with $S_A$. In contrast, the negatively augmented scenarios are drawn from the farthest clusters and represent the least consistent scenario sample.

 


Based on the scenario representation generated by the encoder and derived similarity, we aim to impose the ordering  \vspace{-0.08cm}
\begin{equation}
	{\bf \Gamma}: \quad
	\text{sim}(S_A, S^+) \;\ge\; \text{sim}(S_A, S^*) \;\ge\; \text{sim}(S_A, S^-),
    \label{eq:gamma}
\end{equation} which allows $\text{sim}(S_A, S^*)$ to vary between the bounds of  $\text{sim}(S_A, S^+)$ and $\text{sim}(S_A, S^-)$, rather than enforcing strict consistency where it may not be physically valid.

\subsubsection{Encoder implementation}	
The proposed A$^2$SL framework is agnostic to the choice of encoder architecture, allowing the integration of various existing encoding methods. 
In this study, we employ a bidirectional long-short-term memory (BiLSTM) as the backbone of the scenario model~\cite{graves2005framewise}, due to its effectiveness in explicitly modeling temporal dependencies in environmental ecosystems in both forward and backward directions. 
Let a scenario be
$ 
S = \bigl\{(\pmb{x}_t, \tilde{y}_t)\bigr\}_{t=1}^n.
$ 
At each time step $t$, we construct the concatenated input vector
$ 
[\pmb{x}_t \,;\, \tilde{y}_t],
$ 
which is processed by forward and backward cells to produce hidden states
$\overrightarrow{\pmb{h}}_t$
and
$\overleftarrow{\pmb{h}}_t.$
We form the scenario embedding by taking the mean of the final forward and
backward states:
$\pmb{s}=\frac{\overrightarrow{\pmb{h}}_{n}+\overleftarrow{\pmb{h}}_{1}}{2}
$. The BiLSTM inherently aggregates contextual information from the entire sequence in both temporal directions, which provides a robust scenario representation for subsequent self-supervised similarity learning.

\subsubsection{Multi-level pairwise learning loss}

To effectively capture the similarity order ${\bf \Gamma}$ (Eq.~\ref{eq:gamma}), we  
include multiple pairwise comparisons in the pre-training loss inspired by ranking-based learning methods. Unlike conventional pairwise loss functions that optimize a single similarity ranking at a time, our approach enforces two pairwise constraints simultaneously, allowing the model to learn a more structured similarity hierarchy among scenario augmentations. 

Given the scenario embedding $\pmb{s}_A$ for the anchor scenario $S_A$, and the embeddings of  corresponding contrastive scenarios $\pmb{s}^+$, $\pmb{s}^*$, and $\pmb{s}^-$ for the positively augmented, semi-positively augmented, and negatively augmented scenarios, respectively, the loss function is formulated as:  \vspace{-0.08cm}
\begin{align}
	\mathcal{L}_{\text{En}} = -\log \sigma & \big[ \lambda \big( \text{sim}(\pmb{s}_A, \pmb{s}^+) - \text{sim}(\pmb{s}_A, \pmb{s}^*) \big) \nonumber \\ 
	& + (1 - \lambda) \big( \text{sim}(\pmb{s}_A, \pmb{s}^*) - \text{sim}(\pmb{s}_A, \pmb{s}^-) \big) \big],
\end{align} 
where $\sigma(\cdot)$ denotes the Sigmoid function, and $\text{sim}(\cdot)$ computes the cosine similarity between two scenario embeddings. The tradeoff parameter $\lambda \in [0,1]$ balances the two pairwise constraints, allowing flexibility in optimizing similarity rankings.

Here, $\pmb{s}^+$, $\pmb{s}^*$, and $\pmb{s}^-$ represent the BiLSTM-encoded embeddings of $S^+$, $S^*$, and $S^-$, respectively.
These data chunks are sampled according to predefined rules. During this process, the anchor scenario $S_A$ is explicitly excluded from the candidate pool, guaranteeing it is never re-selected as a positive, semi-positive, or negative example. 

The multi-level pairwise learning loss refines the scenario representations by enforcing structured similarity relationships. Unlike contrastive losses that impose absolute constraints, it maintains relative similarities, ensuring that physically consistent scenarios remain closer in the learned space. This improves retrieval-based augmentation by guiding the encoder to focus on meaningful spatiotemporal relationships. 

\subsection{Retrieval-based Decoder}
\subsubsection{Training the decoder}

The decoder is trained jointly with the scenario encoder, ensuring simultaneous updates to both the scenario embeddings and model parameters. This joint optimization allows the embeddings to capture scenario similarities effectively while improving predictive performance through retrieval-based augmentation. 
For an anchor scenario $S_A$, we compute its similarity with all scenarios in the dataset. The $k$ most similar scenarios are selected and, together with $S_A$, are used as training inputs to a base predictive model.

Let $\mathbb{S}$ be the set of all scenarios and
$\pmb{s}_i$ the embedding of $S_i\!\in\!\mathbb{S}$.
We first obtain the index set of the $k$ scenarios
most similar to the anchor~$S_A$ (excluding itself):  \vspace{-0.08cm}
\begin{equation} 
	\mathcal{I}_k
	= \operatorname{Top}_k\!\Bigl(
	\{\, i \mid S_i\in\mathbb{S},\; i\neq A \,\},
	\; \text{sim}(\pmb{s}_A,\pmb{s}_i)
	\Bigr).
	\label{eq:topk}
\end{equation}
The corresponding retrieval set and the final input batch are  \vspace{-0.08cm}
\begin{equation}
	\mathcal{S}_{\text{re}}
	= \bigl\{\, S_i \mid i\in\mathcal{I}_k \bigr\},
	\qquad
	\mathcal{S}_{\text{in}}
	= \mathcal{S}_{\text{re}}\,\cup\,\{S_A\},
\end{equation} so that $\mathcal{S}_{\text{in}}$ always contains
exactly $k\!+\!1$ \emph{distinct} scenarios.
Each scenario in the batch is fed to the base predictor
$\mathcal{M}_{\alpha}$ to generate prediction: 
\vspace{-0.08cm}
\begin{equation}
	\hat{y}_{t}^{(i)}
	= \mathcal{M}_{\alpha}(S_i,t),
	\qquad
	S_i \in \mathcal{S}_{\text{in}},
\end{equation} where $\hat{y}_{t}^{(i)}$ denotes the predicted target variable
for scenario~$S_i$ at time step~$t$. If fewer than two retrieved chunks contain observed targets, we incrementally increase $k$ until the set includes at least two such chunks.

\subsubsection{Decoder implementation}

The decoder is designed to be model-agnostic so that different network backbones can be plugged in seamlessly.  
In this study, we adopt a long short-term memory (LSTM) network as the default backbone, given its strong performance on sparse hydrological and aquatic time series~\cite{hochreiter1997long}; several alternative architectures are benchmarked in our experiments.
We fix the sequence length at one month, although many days lack ground-truth observations.  
Let $m_{t}^{(i)}\!\in\!\{0,1\}$ be an observation mask that equals $1$ when the target variable for scenario $S_i$ is observed at day~$t$ and $0$ otherwise.  
For every scenario in the input batch $\mathcal{S}_{\text{in}}$ we still generate daily predictions $\hat{y}_{t}^{(i)}$, but only observed days ($m_{t}^{(i)}=1$) contribute to the loss.
To emphasise more relevant scenarios, each scenario’s contribution is further weighted by its cosine similarity to the anchor: \vspace{-0.15cm}
\begin{equation} 
	w_i \;=\;\frac{\text{sim}\bigl(\pmb{s}_i,\pmb{s}_A\bigr)}
	{\sum_{S_j\in\mathcal{S}_{\text{in}}}\text{sim}\bigl(\pmb{s}_j,\pmb{s}_A\bigr)}.
\end{equation}
The resulting similarity-weighted, masked mean-squared-error (MSE) objective is \vspace{-0.15cm}
\begin{equation}
	\label{eq:masked_mse}
	\mathcal{L}_{\text{De}}
	=\;
	\sum_{S_A}\sum_{S_i\in\mathcal{S}_{\text{in}}}
	w_i\;
	\frac{1}{\sum_{t=1}^{n} m_{t}^{(i)}}
	\sum_{t=1}^{n} m_{t}^{(i)}
	\bigl(\hat{y}_{t}^{(i)} - y_{t}^{(i)}\bigr)^{2},
\end{equation}
where $\mathcal{S}_{\text{in}}$ comprises the anchor scenario and its $k$ retrieved neighbors.  
Because the weights $\{w_i\}$ sum to~1, the objective remains on the same scale as a conventional MSE while naturally giving greater influence to scenarios that are more similar to the anchor.
Optimising Eq.~\eqref{eq:masked_mse} jointly updates decoder parameters and refines scenario embeddings, ensuring that retrieval-based augmentation is seamlessly integrated into the learning process.

\subsubsection{Tuning the decoder}
After the initial joint training stage (Eq.~\ref{eq:masked_mse}), we obtain 
a fully trained decoder $\mathcal{M}_{\alpha}$ 
that is shared across all scenarios. During the testing phase,  
the decoder 
is further fine-tuned to better adapt to the 
environmental conditions of each target period before generating the final predictions.  
In particular, for each target short-term period, we treat it as the anchor scenario $S_A$ and optimize 
a scenario-specific decoder $\mathcal{M}_{\beta}^{(A)}$ by fine-tuning $\mathcal{M}_{\alpha}$ on the set of retrieved scenarios most relevant to $S_A$.

During fine-tuning, only the parameters of $\mathcal{M}_{\beta}^{(A)}$ are updated, while the scenario embeddings remain fixed.  
Freezing the embeddings preserves the similarity structure learned during joint training and prevents catastrophic forgetting, thereby enabling the decoder to focus on localized patterns that are specific to $S_A$.  
Once this adaptation is complete, the customized decoder produces the final prediction for $S_A$ as
$
\hat{y}_{t}^{(A)} \;=\; \mathcal{M}_{\beta}^{(A)}\bigl(S_A,\, t\bigr).
$

This two-stage procedure, which first performs global training and then applies scenario-specific fine-tuning, achieves a balance between generalization and adaptability.  
The fine-tuned decoder is especially useful for short-term forecasts, such as monthly predictions under atypical or extreme environmental conditions, where data augmentation provides significant gains in predictive accuracy.

\subsection{Augmentation-Adaptive Mechanism}

Environmental scenarios vary widely in their stability. Some exhibit stable, consistent long-term patterns, whereas others experience significant variability, especially under extreme conditions. Traditional predictive models often perform poorly in data-scarce variable scenarios, where augmentation can help improve accuracy. However, indiscriminate augmentation can introduce noise into stable scenarios, degrading performance.

To balance these trade-offs, we propose an \textit{augmentation-adaptive mechanism} that dynamically selects the most appropriate predictor for each scenario. Specifically, we maintain two families of prediction networks: (i) a \emph{yearly model} $\mathcal{M}_{\gamma}$, which ingests 360-day sequences and is trained independently of the scenario encoder to capture slow dynamics in stable scenarios; and (ii) \emph{monthly models}, where the backbone $\mathcal{M}_{\alpha}$ operates on 30-day chunks, is jointly optimized with the scenario encoder, and serves as the common initialization for fine-tuning. For each variable scenario, we derive a scenario-specific decoder $\mathcal{M}_{\beta}^{(i)}$ (written as $\mathcal{M}_{\beta}$ when the index is clear) by fine-tuning $\mathcal{M}_{\alpha}$ on the scenario itself and its top-$k$ retrieved neighbors. This adaptive switching enables a balanced augmentation strategy that enhances performance without compromising stability.

Let $\mathbb{S}$ be the full scenario set, and $m_t^{(i)}\in\{0,1\}$ be the observation mask introduced in Eq.~\eqref{eq:masked_mse}.  
For each scenario $S_i$ with $n$ days, we compute the masked MSE using the yearly model $\mathcal{M}_{\gamma}$ and fine-tuned monthly model $\mathcal{M}_{\beta}$: \vspace{-0.15cm}
\begin{align}
	\mathcal{E}_{\gamma}(S_i) &= 
	\frac{1}{\sum_{t=1}^{n} m_{t}^{(i)}} 
	\sum_{t=1}^{n} m_{t}^{(i)}
	\bigl(\mathcal{M}_{\gamma}(S_i,t)-y_t^{(i)}\bigr)^{2},\\
	\mathcal{E}_{\beta}(S_i)  &= 
	\frac{1}{\sum_{t=1}^{n} m_{t}^{(i)}} 
	\sum_{t=1}^{n} m_{t}^{(i)}
	\bigl(\mathcal{M}_{\beta}(S_i,t)-y_t^{(i)}\bigr)^{2}.
\end{align}
A scenario is labeled as \emph{variable} if the fine-tuned model yields a lower MSE: \vspace{-0.08cm}
\begin{equation}
	\mathbb{P}=\bigl\{S_i\mid\mathcal{E}_{\beta}(S_i)<\mathcal{E}_{\gamma}(S_i)\bigr\},\qquad
	\mathbb{N}=\mathbb{S}\setminus\mathbb{P},
\end{equation} 
where $\mathbb{P}$ and $\mathbb{N}$ denote the variable and stable subsets. 

To automate this classification, we introduce a discriminative model $D_{\omega}(S_i)$, parameterized by $\omega$, which outputs a probability in $[0, 1]$ indicating the likelihood of scenario $S_i$ being variable. We use a predefined threshold $\tau \in (0,1)$ to convert this probability into a binary decision. During training, the discriminator is optimized by maximizing the standard binary cross-entropy objective: \vspace{-0.08cm}
\begin{equation}
	\max_{\omega}\;\;  
	\mathbb{E}_{S_i \in \mathbb{P}}\left[\log D_{\omega}(S_i)\right] 
	+ 
	\mathbb{E}_{S_i \in \mathbb{N}}\left[\log\left(1 - D_{\omega}(S_i)\right)\right].
	\label{eq:discriminator}
\end{equation}

The discriminator is trained to converge on a held-out validation set. For a new scenario $S_i$, we compute $p=D_{\omega}(S_i)$.  
The final prediction for each scenario is generated by exactly one model: \vspace{-0.08cm}
\begin{equation}
	\hat{y}_t =
	\begin{cases}
		\mathcal{M}_{\gamma}(S_i,t), & p \leq \tau \\[3pt]
		\mathcal{M}_{\beta}(S_i,t),  & p > \tau.
	\end{cases}
\end{equation}

At inference time, the augmentation-adaptive mechanism ensures that retrieval-based augmentation is applied only when beneficial, thereby enhancing accuracy in variable scenarios while maintaining robust performance in stable scenarios.

\section{Experimental Evaluation}
\label{sec:exp}

We conduct extensive experiments across various lakes in the Midwestern United States to address research questions:

\begin{itemize}
	\item \textbf{RQ1.} How does the proposed A$^2$SL framework compare to baseline methods in predictive performance?
	\item \textbf{RQ2.} Can A$^2$SL serve as a general data augmentation strategy to enhance models under atypical conditions?
	\item \textbf{RQ3.} How well does A$^2$SL capture temporal patterns in water temperature and DO concentration time series?
	\item \textbf{RQ4.} What types of data chunks does A$^2$SL retrieve for a given anchor scenario?
\end{itemize}

\subsection{Data Preparation}

\begin{figure}
	\centering
	\centerline{\includegraphics[width=0.72\linewidth]{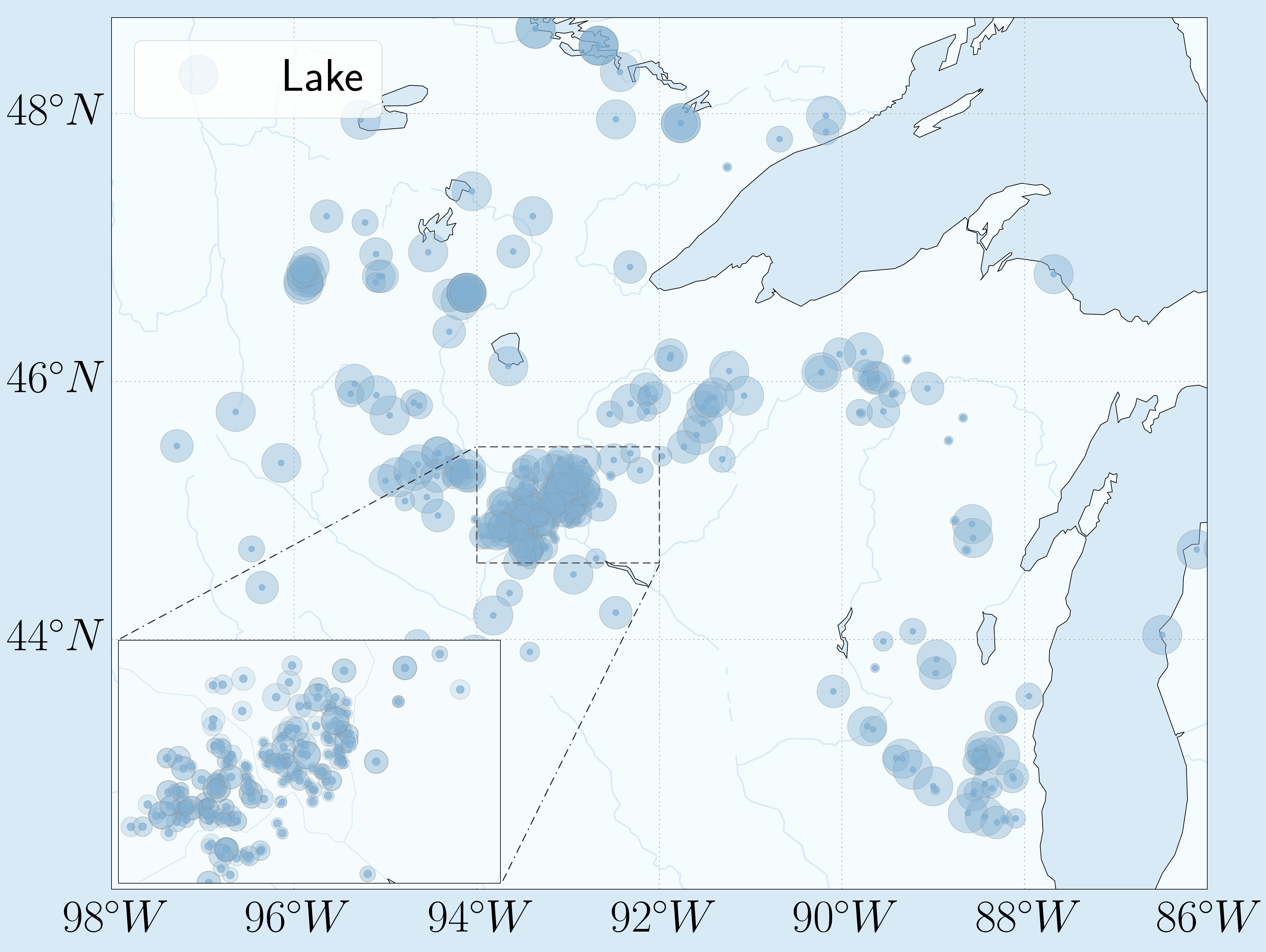}}
	\caption{Map of 356 tested lakes. } 
	\label{fig:3}
\end{figure}

We evaluate the proposed A$^2$SL framework for predicting water temperature and DO concentrations using a comprehensive dataset spanning 41 years (1979–2019). The dataset comprises ecological observations from 356 lakes across the Midwestern USA, as illustrated in Figure~\ref{fig:3}. In the figure, color intensity represents lake depth, while marker size corresponds to surface area. 
The dataset contains approximately 1.75 million daily records, each including 47 environmental features such as morphometric attributes, meteorological conditions, trophic states, and land use characteristics. Data sources are described in~\cite{meyer2024national,yu2024adaptive,willard2021predicting}. Specifically, water temperature data were obtained from the U.S. Geological Survey~\cite{willard2021predicting}, with 476,215 observed measurements recorded across depths on 57,156 distinct days. DO concentration data were retrieved from the Water Quality Portal (WQP), encompassing 23,192 days of observations. Lake residence time was sourced from the HydroLAKES dataset. Trophic state probabilities were derived from~\cite{meyer2024national}. Land use proportions within each lake's watershed were extracted from the National Land Cover Database (NLCD). For model development, the dataset is segmented as follows: data collected up to 2011 form the training set, data spanning 2012 to 2015 constitute the validation set, and data from 2016 to 2019 are designated as the test set.

\newcolumntype{L}[1]{>{\raggedright\arraybackslash}p{#1}}
\newcolumntype{C}[1]{>{\centering\arraybackslash}p{#1}}
\newcolumntype{R}[1]{>{\raggedleft\arraybackslash}p{#1}}

\begin{table*}[t]
	\centering
	\setlength{\tabcolsep}{6pt}      
	\renewcommand{\arraystretch}{1.12}
	\caption{Comparative RMSE performance.  Lower values are better; grey numbers denote standard deviation.}
		\vspace{-0.1cm}
	\label{tab:rmse_results}
	\begin{tabular}{L{2.7cm}C{2.0cm}C{2.0cm}C{2.0cm}C{2.0cm}C{2.0cm}C{2.0cm}}
		\toprule
		\multirow{2}{*}{Algorithm} &
		\multicolumn{3}{c}{\textbf{Water temperature ($^\circ$C)}} &
		\multicolumn{3}{c}{\textbf{DO concentration ($ \bf{g / m^{3}}$)}} \\
		\cmidrule(lr){2-4} \cmidrule(lr){5-7}
		& Summer (epi.) & Summer (hyp.) & Fall to spring & Summer (epi.) & Summer (hyp.) & Fall to spring \\
		\midrule
		Physics-based & 1.608\,\textcolor{gray}{(0.000)} & 2.544\,\textcolor{gray}{(0.000)} & 1.665\,\textcolor{gray}{(0.000)} & 1.954\,\textcolor{gray}{(0.000)} & 3.037\,\textcolor{gray}{(0.000)} & 3.365\,\textcolor{gray}{(0.000)} \\
		LSTM          & 1.330\,\textcolor{gray}{(0.043)} & 1.815\,\textcolor{gray}{(0.038)} & 1.039\,\textcolor{gray}{(0.047)} & 1.943\,\textcolor{gray}{(0.023)} & 3.460\,\textcolor{gray}{(0.204)} & 2.474\,\textcolor{gray}{(0.029)} \\
		EA-LSTM       & 1.617\,\textcolor{gray}{(0.226)} & 2.778\,\textcolor{gray}{(0.386)} & 1.266\,\textcolor{gray}{(0.173)} & 1.969\,\textcolor{gray}{(0.014)} & 2.883\,\textcolor{gray}{(0.019)} & 2.580\,\textcolor{gray}{(0.045)} \\
		PT-LSTM       & 1.295\,\textcolor{gray}{(0.013)} & 1.726\,\textcolor{gray}{(0.014)} & 1.038\,\textcolor{gray}{(0.069)} & 1.896\,\textcolor{gray}{(0.029)} & 2.830\,\textcolor{gray}{(0.038)} & 2.398\,\textcolor{gray}{(0.145)} \\
		iTransformer  & 1.841\,\textcolor{gray}{(0.072)} & 2.181\,\textcolor{gray}{(0.073)} & 2.088\,\textcolor{gray}{(0.087)} & 2.020\,\textcolor{gray}{(0.064)} & 3.044\,\textcolor{gray}{(0.059)} & 2.435\,\textcolor{gray}{(0.059)} \\
		TSMixer       & 1.753\,\textcolor{gray}{(0.283)} & 1.676\,\textcolor{gray}{(0.110)} & 1.533\,\textcolor{gray}{(0.331)} & 2.058\,\textcolor{gray}{(0.050)} & 2.947\,\textcolor{gray}{(0.014)} & 2.519\,\textcolor{gray}{(0.060)} \\
		TimesNet      & 1.812\,\textcolor{gray}{(0.061)} & 1.723\,\textcolor{gray}{(0.031)} & 2.140\,\textcolor{gray}{(0.252)} & 1.966\,\textcolor{gray}{(0.023)} & 2.923\,\textcolor{gray}{(0.047)} & 2.445\,\textcolor{gray}{(0.099)} \\
		TimeMixer     & 2.348\,\textcolor{gray}{(0.252)} & 2.750\,\textcolor{gray}{(0.732)} & 2.454\,\textcolor{gray}{(0.551)} & 2.186\,\textcolor{gray}{(0.228)} & 3.846\,\textcolor{gray}{(0.836)} & 2.606\,\textcolor{gray}{(0.137)} \\
		\textbf{A$^{2}$SL } & \textbf{1.253}\,\textcolor{gray}{(0.023)} & \textbf{1.604}\,\textcolor{gray}{(0.019)} & \textbf{0.967}\,\textcolor{gray}{(0.033)} & \textbf{1.757}\,\textcolor{gray}{(0.013)} & \textbf{2.719}\,\textcolor{gray}{(0.036)} & \textbf{2.397}\,\textcolor{gray}{(0.120)} \\
		\bottomrule
	\end{tabular}
\end{table*}

\begin{figure*}[t]
	\centering
			\vspace{-0.15cm}
	\includegraphics[width=0.94\linewidth]{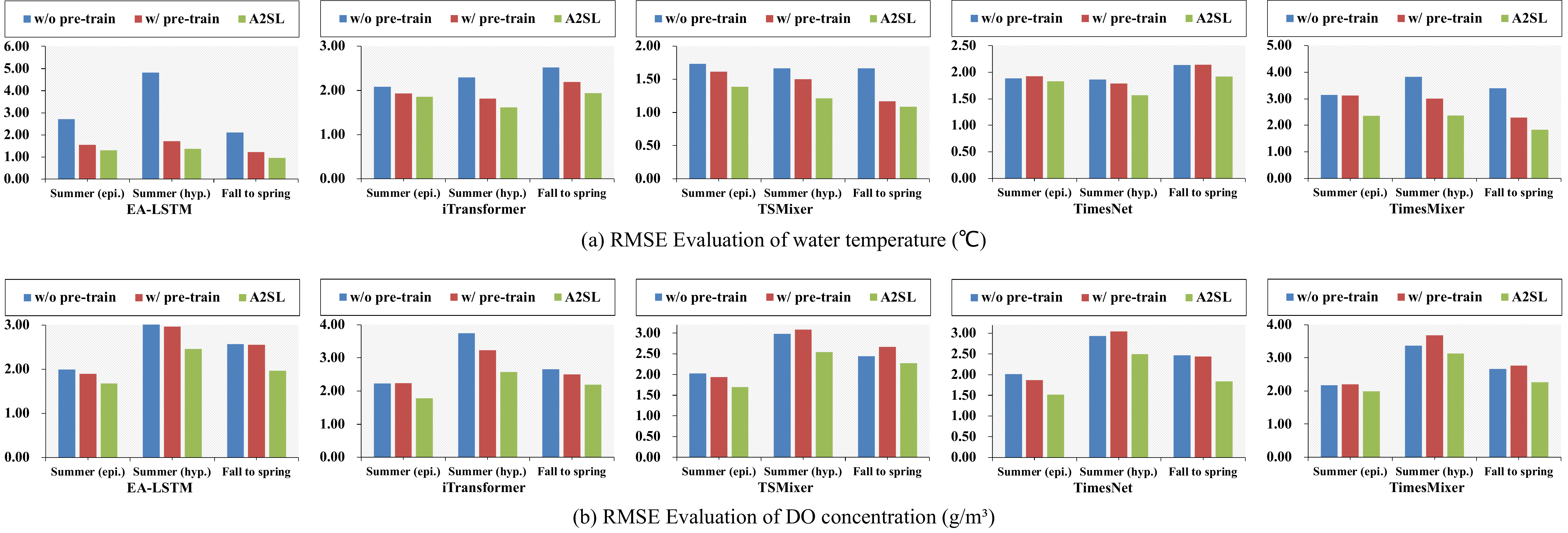}
	\vspace{-0.25cm}
	\caption{Comparison under three training settings: (1) without pre-training, (2) with simulated-label pre-training, and (3) using our proposed A$^2$SL framework. } 
	\vspace{-0.4cm}
	\label{fig:combined_bar}
\end{figure*}

\subsection{Baselines} 
To evaluate the effectiveness of our A$^2$SL framework, we compare it against several baseline models commonly used in time series forecasting:

\begin{itemize}
	\item \textbf{Physics-based}~\cite{hipsey2019general,ladwig2022long}: As discussed in Section~\ref{sec:3A}, these baselines represent state-of-the-art physics-based models. The flux features used by these models are obtained through calibration with observed data.
	\item \textbf{LSTM}~\cite{hochreiter1997long}: A recurrent neural network architecture designed to capture long-term dependencies in sequential data, widely used for time series forecasting.
	\item \textbf{EA-LSTM}~\cite{kratzert2019towards}: An Entity-Aware LSTM network that integrates hydrological behavior and distinguishes between similar dynamic behaviors across various entities, including distinct morphometric characteristics.
	\item \textbf{PT-LSTM}~\cite{yu2021pt}: An enhanced LSTM model that incorporates time attributes via position encoding and a dedicated time gate, enabling better capture of temporal patterns. 
	\item \textbf{iTransformer}~\cite{liuitransformer}: A Transformer-based model designed for irregular time series data, incorporating adaptive attention mechanisms to better capture temporal dependencies.
	\item \textbf{TSMixer}~\cite{chen2023tsmixer}: An all-MLP architecture for time series forecasting that integrates historical data, future known inputs, and static contextual information through conditional feature mixing and mixer layers.
	\item \textbf{TimesNet}~\cite{wutimesnet}: A framework that transforms one-dimensional time series into two-dimensional representations, allowing the use of 2D convolutional kernels to extract temporal patterns.
	\item \textbf{TimeMixer}~\cite{wangtimemixer}: A model that explores mixing-based architectures for time series forecasting, focusing on effectively capturing short- and long-term dependencies.
\end{itemize}

\noindent Among these baselines, we designate LSTM (w/o pre-training) as a baseline, trained solely on observed labels within the training set. In contrast, all other baseline models undergo pre-training on simulated labels before being fine-tuned on observed data. This pre-training step is essential, as these models are highly data-dependent, and their performance degrades significantly without sufficient pre-training. To assess the effectiveness of A$^2$SL, we use LSTM as the base predictive model in the main evaluation. Additionally, we examine A$^2$SL as a general data augmentation strategy by applying it to other advanced models, evaluating its ability to enhance predictive accuracy under atypical or extreme environmental conditions. 
Hyperparameters for all methods were carefully tuned via an extensive grid search. Optimal configurations were selected based on validation performance. We explored batch sizes ranging from 8 to 32, learning rates from 0.001 to 0.05, and hidden layer dimensions from 20 to 200. For Transformer-based models, the number of attention heads was varied between 4 and 16. For A$^2$SL specifically, we also tuned the trade-off parameter $\lambda \in \{0.0, 0.1, \dots, 1.0\}$ and the discriminator threshold $\tau \in \{0.0, 0.1, \dots, 1.0\}$. 

\subsection{Experimental Results} 

\subsubsection{Performance comparison}

 Table~\ref{tab:rmse_results} compares the performance of A$^{2}$SL integrated with a standard LSTM backbone against baselines. It measures performance based on the root mean square
error (RMSE), including both mean and standard deviation, calculated over five runs.
Key findings from this analysis are summarized below (RQ1):

\begin{enumerate}[label=(\alph*)]
	\item	Physics-based models perform worse than most data-driven models across the majority of tasks, reflecting its limitations in capturing detailed thermal and biochemical variability, particularly within hypolimnion layers.
	\item 
	A basic LSTM significantly outperforms the physics-based baseline. Enhanced models as EA-LSTM and PT-LSTM could offer further incremental improvements, particularly noticeable in hypolimnion DO predictions.
	\item 
	iTransformer and Mixer-style models are competitive for temperature prediction tasks but noticeably weaker for DO concentration predictions. This weakness is likely due to their sensitivity to data sparsity.
	\item 
	With identical LSTM capacity, A$^{2}$SL consistently achieves the lowest RMSE across all six prediction tasks. In the challenging summer hypolimnion setting, A$^{2}$SL achieves an 11.6\% RMSE reduction in water temperature prediction and a 21.4\% reduction in DO concentration compared to LSTM. The improvement reaches up to 42.3\% and 29.3\% when compared against EA-LSTM and TimeMixer, respectively.
	
\end{enumerate}

To assess the broader applicability of A$^{2}$SL, we further integrate it with five predictive models and evaluate its effectiveness under atypical environmental conditions (Fig.~\ref{fig:combined_bar}). Key observations from this extended analysis include (RQ2):

\begin{enumerate}[label=(\alph*)]
	\item
	Pre-training with simulated labels consistently outperforms direct supervised training across nearly all tested models and tasks. This result underscores the effectiveness of weak supervision in data-scarce scenarios.
	\item 
	A$^{2}$SL consistently achieves the lowest RMSE across all tasks and demonstrates robust performance gains for both temperature and DO predictions, applicable across diverse model architectures. These results emphasize A$^{2}$SL as a robust, model-agnostic augmentation strategy.
    
\end{enumerate}

\begin{figure}
		\centering
		\includegraphics[width=0.9\linewidth]{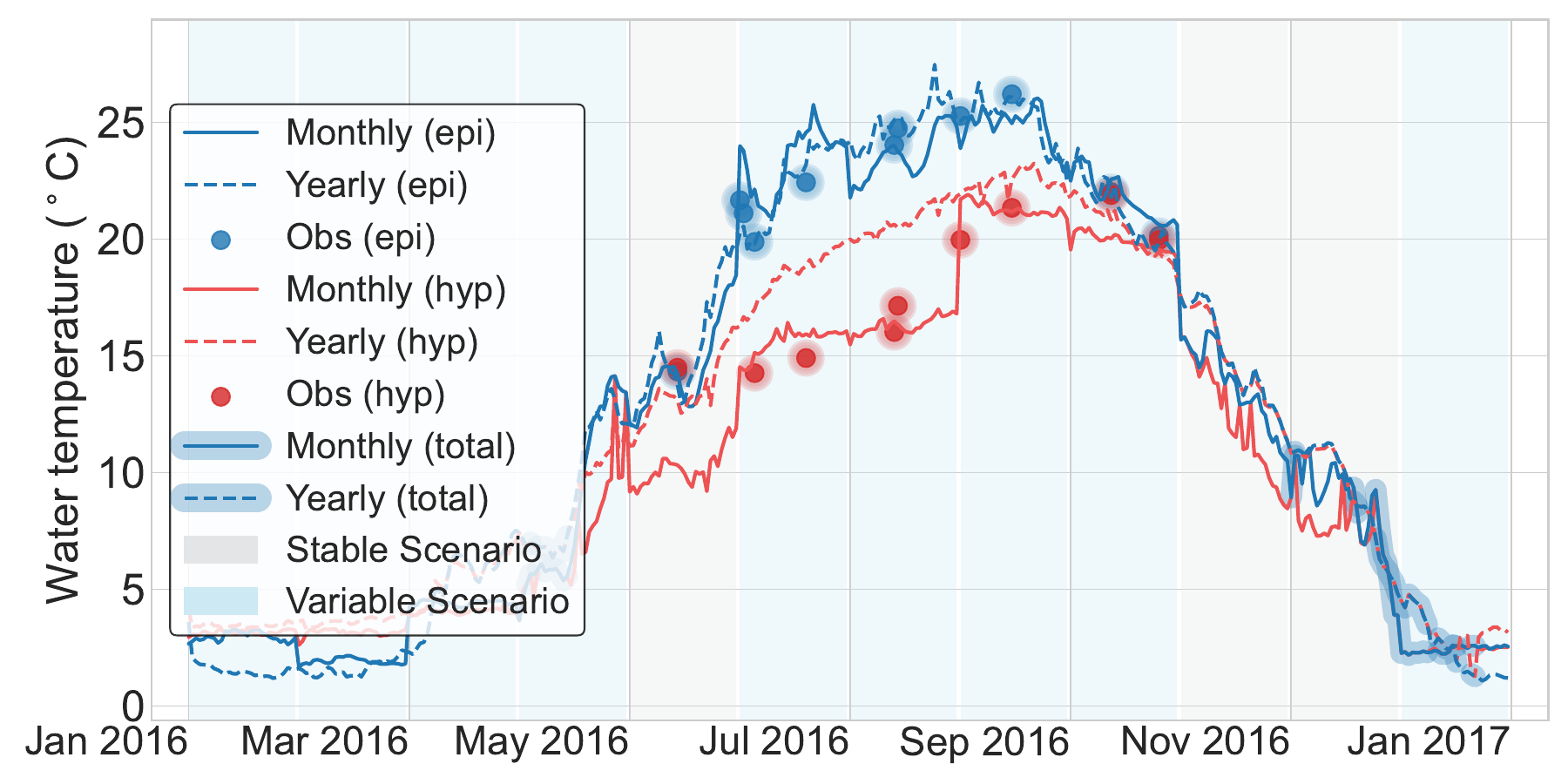}
		\vspace{-0.3cm}
		\caption{Water temperature predictions with dynamic switching by A$^2$SL.} 
		\label{fig:watertimeseries}
	\end{figure}

	\begin{figure}
		\centering
		\includegraphics[width=0.9\linewidth]{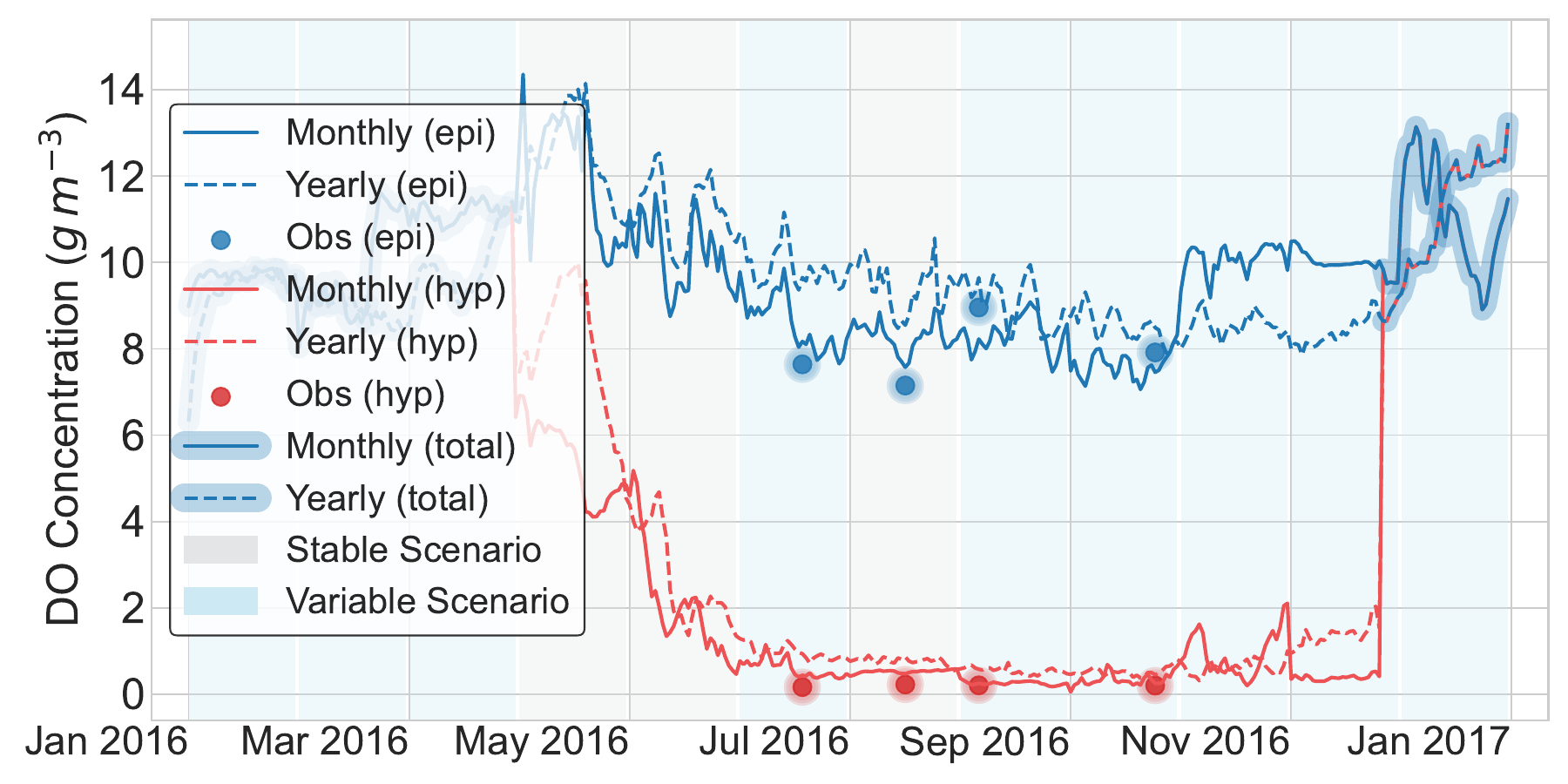}
		\vspace{-0.3cm}
		\caption{DO concentration predictions with dynamic switching by A$^2$SL. } 
		\label{fig:DOtimeseries}
	\end{figure}

\subsubsection{Time-series analysis (RQ3)}

Fig.~\ref{fig:watertimeseries} illustrates one year of water temperature dynamics in a representative lake, while Fig.~\ref{fig:DOtimeseries} presents the corresponding DO concentration series. Both figures utilize background shading to indicate the augmentation-adaptive mechanism's model choice: grey intervals represent stable periods handled by the yearly model, whereas light-blue intervals indicate variable conditions, prompting dynamic switching to the monthly model. 
Several key insights emerge from this analysis.

During rapid warming events, such as those observed between August and September, the augmentation-adaptive mechanism correctly identifies high variability. The benefit of dynamic model switching is particularly pronounced for hypolimnetic conditions. By switching to the monthly model, the predictions closely track the short-term temperature peaks, which the yearly model tends to smooth out and underestimate. 
During periods when oxygen-rich waters rapidly decrease, the yearly model consistently underestimates the decline rate. In contrast, the monthly model, activated selectively during variable intervals, accurately captures the step-like decreases and closely aligns with the sparse observational data.

Overall, these results underscore the effectiveness of A$^{2}$SL. The adaptive selection between yearly and monthly models significantly enhances the ability to capture short-term fluctuations and critical transitions in both water temperature and DO concentrations, thus providing a robust predictive advantage under dynamically changing environmental conditions.

	\begin{figure}
		\centering
		\includegraphics[width=0.96\linewidth]{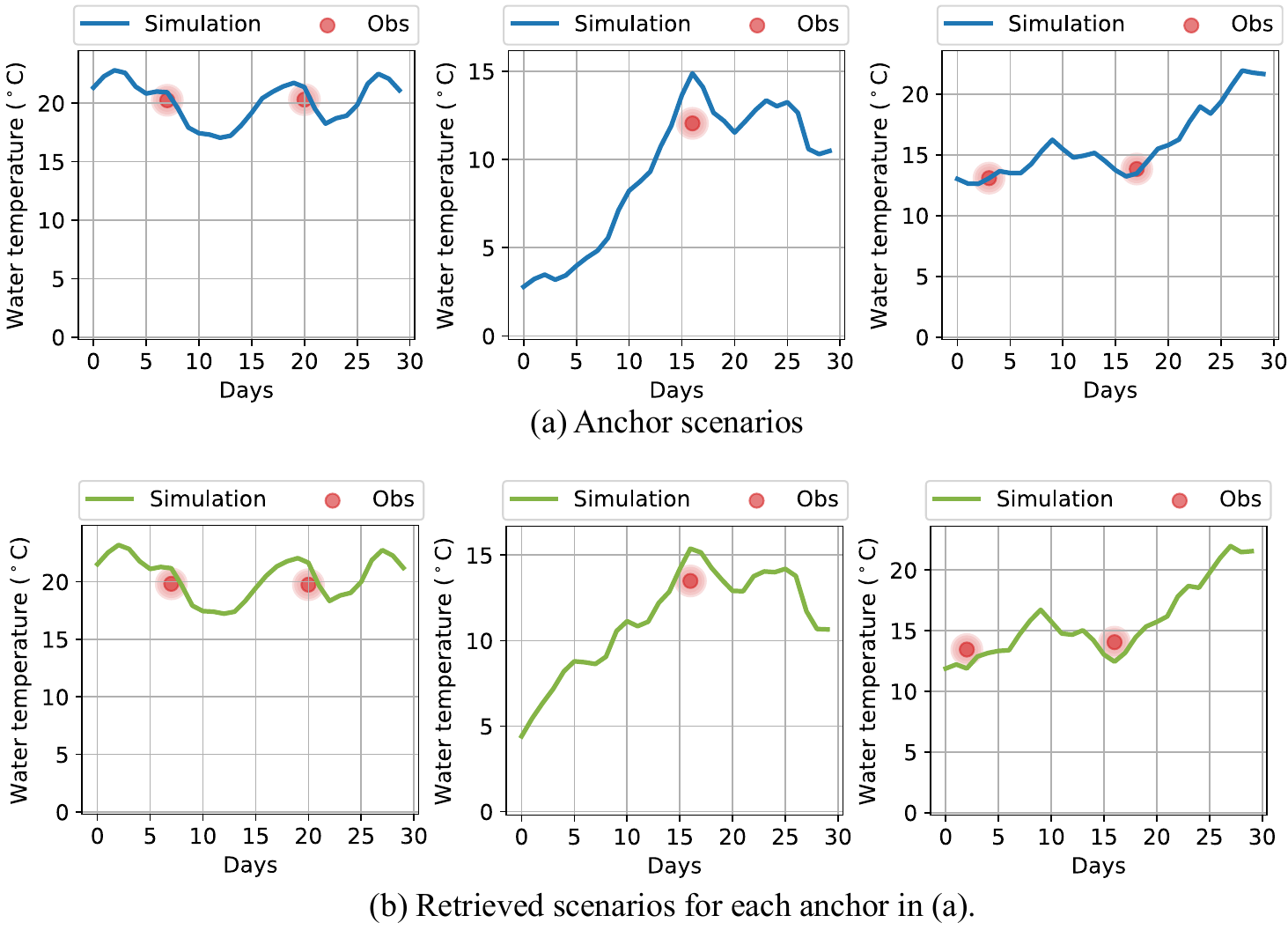}
		\vspace{-0.4cm}
		\caption{A$^{2}$SL retrieval for water temperature predictions. } 
		\label{fig:water_simulation}
	\end{figure}
    
	\begin{figure}
		\centering
		\includegraphics[width=0.96\linewidth]{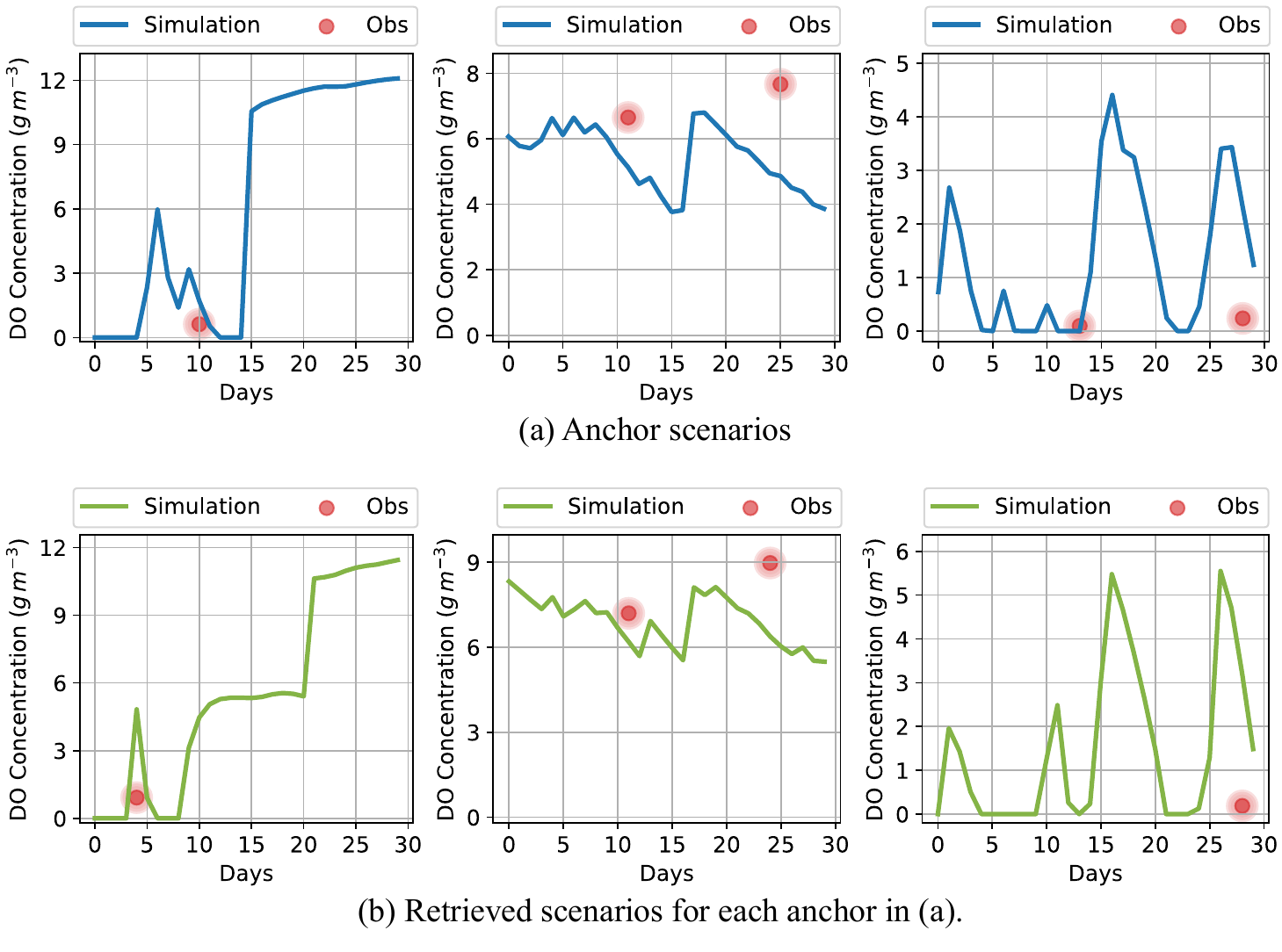}
		\vspace{-0.4cm}
		\caption{A$^{2}$SL retrieval for DO concentration predictions. } 
		\label{fig:DO_simulation}
	\end{figure}

\subsubsection{Retrieval analysis (RQ4)}

Fig.~\ref{fig:water_simulation} and Fig.~\ref{fig:DO_simulation} illustrate the data-chunk retrieval process in A$^{2}$SL. The top row in each figure presents three representative anchor scenarios (one-month windows) from the test set, shown as process-based simulations overlaid with sparse observations. The bottom row shows the corresponding nearest-neighbor scenarios retrieved by the encoder from different lakes or years.

Across both variables, the retrieved chunks exhibit strong temporal resemblance to their respective anchors. Rising limbs align with rising limbs, double-peak patterns are matched with similar structures, and sharp step-like drops are paired with equally abrupt transitions. This demonstrates that A$^{2}$SL prioritizes short-term temporal dynamics when computing scenario similarity, rather than relying solely on raw magnitudes.
In addition to shape similarity, the retrieved scenarios often include observations at time steps where the anchor is unobserved, effectively densifying the supervision signal during fine-tuning. These label-enriching and structurally aligned retrievals play a key role in boosting predictive accuracy, particularly under sparse or highly variable conditions, and explain the consistent performance gains observed across all tasks and model architectures.

\subsubsection{Discriminator threshold sensitivity test} 

\begin{figure}
		\centering
		\includegraphics[width=0.9\linewidth]{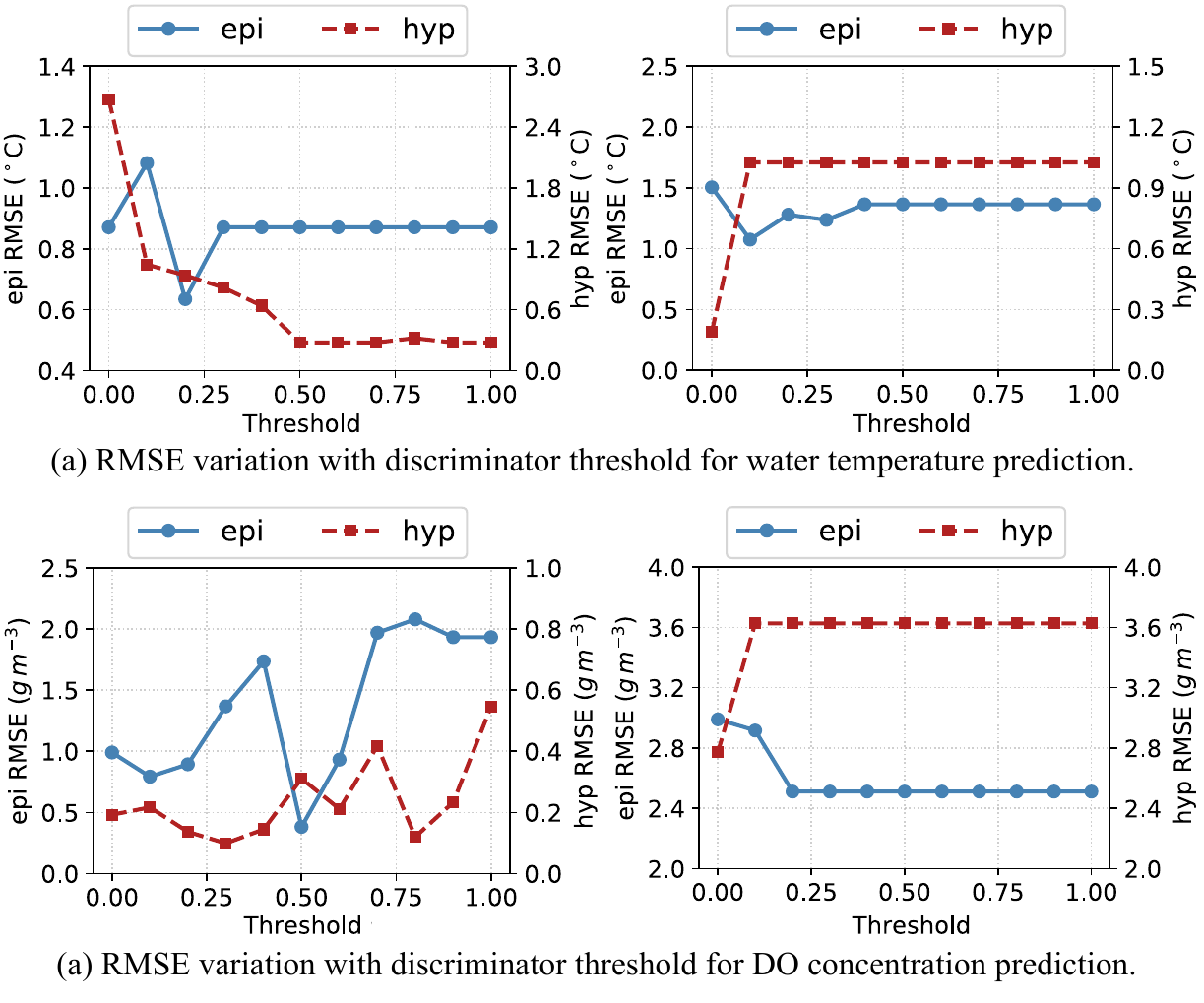}
		\vspace{-0.3cm}
		\caption{Sensitivity of prediction performance to the discriminator threshold.} 
		\label{fig:threshold}
	\end{figure}

To assess the effectiveness of the augmentation-adaptive mechanism in A$^2$SL, we analyze how prediction accuracy varies with changes in the discriminator threshold $\tau$. As shown in Fig.~\ref{fig:threshold}, each curve represents the RMSE performance for a specific lake-year pair under different values of $\tau$, which controls the dynamic switch between the yearly and monthly models. We observe that no universal threshold yields optimal performance across all scenarios. In some cases, lower values of $\tau$ (i.e., favoring the monthly model) produce better accuracy, while others perform better at higher $\tau$ values, indicating a preference for the yearly model. This variability highlights the importance of a learned, scenario-specific discriminator rather than a static threshold. Additionally, many curves exhibit flat RMSE regions beyond certain thresholds—either at the lower or upper end—indicating consistent model preference. For instance, a plateau beginning at $\tau=0$ implies that the monthly model is generally superior for that lake and year, while a flat region at higher $\tau$ suggests the yearly model suffices. Some curves also display U-shaped behavior, where RMSE first decreases and then increases across the $\tau$ range, implying that adaptive switching between models outperforms relying on either alone. 


\section{Related Work}
	\label{sec:related_work}
	
\textit{\textbf{Process-based modeling.}}
Process-based models, also referred to as physics-based or theory-driven models, have long served as the foundation for environmental system modeling~\cite{yu2025survey,yu2025foundation}. These models are built upon first principles from physics, chemistry, and biology, using differential equations to represent key processes such as energy fluxes, hydrological transport, and biogeochemical cycling. They have been widely adopted in domains such as watershed hydrology~\cite{markstrom2012p2s}, lake thermal regimes~\cite{hipsey2019general}, and agroecosystem dynamics~\cite{zhou2021quantifying}.
In freshwater systems, vertical one-dimensional aquatic ecosystem models (AEMs) such as GLM-AED~\cite{hipsey2019general}, WET~\cite{nielsen2017open}, and PCLake+\cite{janssen2019pclake+} integrate hydrodynamics with water quality and ecological processes. However, these models typically require extensive calibration, exhibit sensitivity to parameter choices, and suffer from equifinality, hindering robust generalization\cite{janssen2015exploring,luo2018autocalibration}.
Moreover, process-based models are calibrated under stable environmental regimes and struggle to capture short-term variability or rapid transitions~\cite{gupta2014debates}. 

\textit{\textbf{Data-driven and hybrid models.}}
Machine learning has significantly advanced predictive modeling in agriculture~\cite{ghazaryan2020crop,van2020crop}, hydrology~\cite{kratzert2019towards}, and limnology~\cite{rahmani2021exploring,willard2021predicting}. Modern architectures, including recurrent and attention-based networks, often outperform traditional process-based methods when ample observations are available~\cite{kochkov2024neural,gupta2024unified}. However, their performance can degrade under covariate shift or observation sparsity, reflecting their data-intensive and black-box nature~\cite{hey2009fourth}. To address these limitations, hybrid physics–machine learning models have emerged, integrating domain knowledge and physical principles into data-driven workflows. These hybrid paradigms improve generalization and ensure consistency with known ecological or hydrological dynamics by embedding mechanistic constraints~\cite{national2018improving,yu2024adaptive,yu2025physics}. For instance, process-guided neural networks have been shown to yield improved long-term predictions in lake biogeochemistry by respecting stratification principles~\cite{yu2024adaptive}.
A common strategy in data-driven modeling is transfer learning~\cite{willard2021predicting}. In contrast to prior work that focuses solely on parameter transfer, A$^2$SL actively retrieves physically analogous scenarios from simulation data to enrich training batches. 

\textit{\textbf{Self-supervised learning and retrieval augmentation.}}  
Self-supervised learning has become a foundational approach for representation learning in environmental sciences, enabling models to leverage vast amounts of unlabeled data through contrastive, masked-token, or predictive pretext tasks. However, most existing self-supervised approaches adopt a binary similarity formulation, which fails to capture the nuanced and continuous spectrum of physical similarity inherent in geoscientific systems. We address this limitation by proposing a multi-level pairwise learning loss that introduces an intermediate “semi-positive” class, enhancing the physical fidelity of learned embeddings.
Meanwhile, Retrieval-Augmented Generation (RAG) has emerged as a powerful paradigm for overcoming the limitations of static model memory by incorporating dynamic external knowledge at inference time~\cite{lewis2020retrieval,abootorabi2024multimodal,yu2025rag}. Originally developed for language modeling, RAG has since expanded into multimodal and graph-based domains for more contextual and structured retrieval~\cite{peng2024graphrag}. 
The environmental analogue shares conceptual underpinnings~\cite{janssen2015exploring}, but remains largely decoupled from learned neural encoders and self-supervised augmentation pipelines.
Recent advances in curriculum retrieval, difficulty-aware sampling, and regime switching~\cite{yu2024adaptive,ravirathinam2024combining} underscore the need for selective data augmentation. Our proposed framework extends these ideas. By enabling regime-aware augmentation, A$^2$SL improves robustness and generalization across diverse geophysical regimes.

\section{Conclusion} 
	\label{sec:conclustion}
	
In this paper, we presented an \textit{Augmentation-Adaptive Self-Supervised Learning (A$^2$SL)} framework for modeling freshwater ecosystems.
A$^2$SL retrieves physically consistent data chunks based on scenario similarity and dynamically selects between specialized models.
Empirical evaluations of real-world freshwater systems demonstrate consistent gains over both process-based and ML baselines.
As a general-purpose framework for adaptive augmentation, A$^2$SL holds promise beyond freshwater systems, paving the way for scalable and robust environmental modeling across scientific domains.

\section{Acknowledgments}
This work was supported by the National Science Foundation (NSF) grants 2239175, 2316305, 2147195, 2425844, 2425845, 2430978, 2126474, 2530609, 2530610, and 2213549; the USGS awards  G21AC10564 and G22AC00266; the NASA grants 80NSSC24K1061 and 80NSSC25K0013; and the NSF NCAR's Derecho HPC system. This research was also supported in part by the University of Pittsburgh Center for Research Computing through the resources provided.


	\bibliographystyle{IEEEtran}
	\bibliography{mybibliography}

\end{document}